\relax

\documentclass[letterpaper]{article}
\usepackage{aaai20}
\usepackage{times}
\usepackage{helvet}
\usepackage{courier}
\usepackage[hyphens]{url}
\usepackage{graphicx}
\usepackage{amsfonts}
\usepackage{amsmath}
\usepackage{multirow}
\usepackage{booktabs}
\usepackage{graphicx}
\title{Towards Better Forecasting by Fusing \\Near and Distant Future Visions}
\author{Jiezhu Cheng,\textsuperscript{\rm 1,\rm 2} 
Kaizhu Huang,\textsuperscript{\rm 3,\rm 4} 
Zibin Zheng\textsuperscript{\rm 1,\rm 2}\\
\textsuperscript{\rm 1} School of Data and Computer Science, Sun Yat-sen University, Guangzhou, China\\ 
\textsuperscript{\rm 2} National Engineering Research Center of Digital Life, Sun Yat-sen University, Guangzhou, China\\
\textsuperscript{\rm 3} Department of Electrical and Electronic Engineering, Xi’an Jiaotong-Liverpool University, Suzhou, China\\
\textsuperscript{\rm 4} Alibaba-Zhejiang University Joint Institute of Frontier Technologies, Hangzhou, China
\\
chengjzh@mail2.sysu.edu.cn, kaizhu.huang@xjtlu.edu.cn, zibinzheng2@yeah.net
}
\begin{document}
\maketitle
\begin{abstract}
Multivariate time series forecasting is an important yet challenging problem in machine learning. Most existing approaches only forecast the series value of one future moment, ignoring the interactions between predictions of future moments with different temporal distance. Such a deficiency probably prevents the model from getting enough information about the future, thus limiting the forecasting accuracy. To address this problem, we propose \emph{Multi-Level Construal Neural Network} (MLCNN), a novel multi-task deep learning framework. Inspired by the Construal Level Theory of psychology, this model aims to improve the predictive performance by fusing forecasting information (i.e., future visions) of different future time. We first use the Convolution Neural Network to extract multi-level abstract representations of the raw data for near and distant future predictions. We then model the interplay between multiple predictive tasks and fuse their future visions through a modified Encoder-Decoder architecture. Finally, we combine traditional Autoregression model with the neural network to solve the scale insensitive problem. Experiments on three real-world datasets show that our method achieves statistically significant improvements compared to the most state-of-the-art baseline methods, with average 4.59\% reduction on RMSE metric and average 6.87\% reduction on MAE metric.
\end{abstract}

\section{Introduction}

\noindent Multivariate time series consists of experimental data with multiple variables observed at different points in time. They occur everywhere in our daily life, from the energy consumption, the traffic flow, to the stock prices. In such fields, effective decision often requires accurate prediction on relevant time series data. For example, knowing demand for electricity in the next few hours could help us devise a better energy use plan, and forecasting of stock market in the near or distant future could produce more profit.

Multivariate time series forecasting focuses on predicting the future outcomes of each variable given their past. As it is difficult to estimate exact future values, it is generally considered that the future observations can be subject to a conditional probability distribution of the past observations. In this case,  the conditional expectation of the distribution can be given as a function of the past values:
\begin{equation}
  \mathbb{E}[X_{t+h}|X_t, ..., X_{t-p+1}] = f(X_t, ..., X_{t-p+1})~. \label{equ:MTSDef}
\end{equation}
For simplicity, we use $X_{t+h}$ to represent the conditional mean $\mathbb{E}[X_{t+h}|X_t, ..., X_{t-p+1}]$ in later descriptions.

\begin{figure}[htbp]
  \centering
  \includegraphics[width=.95\columnwidth]{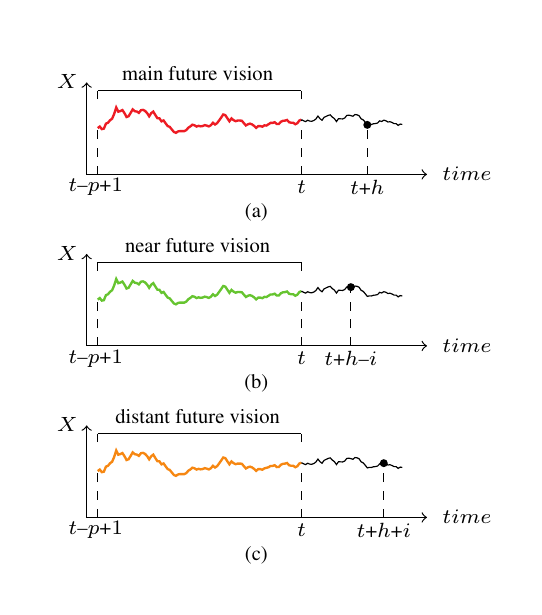}
  \caption{(a) Main predictive task on $X_{t+h}$. (b) Auxiliary predictive task on $X_{t+h-i}$ with near future vision. (c) Auxiliary predictive task on $X_{t+h+i}$ with distant future vision.}
  \label{fig:future_visions}
\end{figure}

Researchers have been studying the forecasting problem~(\ref{equ:MTSDef}) for years, developing all kinds of linear, non-linear and hybrid models for better predictions~\cite{TS08,Survey02}. However, given the past values, most of these models only estimate the conditional mean at the future moment $X_{t+h}$ or in a continuous future window $\{X_{t+1}, ..., X_{t+h}\}$, using a single model architecture without considering the link between predictions of different future moments. This drawback may limit the models' generalization ability, since only one kind of vision about the future is obtained.  Figure~\ref{fig:future_visions} shows three predictive tasks on $X_{t+h}, X_{t+h-i}$ and $X_{t+h+i}$, where $0<i<h$. Although they are performed based on the same past observations, different temporal distances from future observations give distinct future vision to each task. Current forecasting methods such as AR~\cite{TS02}, AECRNN~\cite{AECRNN}, DARNN~\cite{DARNN} and LSTNet~\cite{LSTNet} perform these tasks independently with a single model architecture, ignoring the interplay between them. To our knowledge, there are few methods that model the interactions between multiple predictive tasks and fuse their future visions to improve the main task.

In this paper, we investigate whether the fusion of near and distant future visions could improve the performance of the main predictive task, as shown in Figure~\ref{fig:future_visions}. Inspired by the Construal Level Theory (CLT)~\cite{CLT01} revealing that people use different levels of abstract construals to predict future events, we propose a novel multi-task deep learning framework called \emph{Multi-Level Construal Neural Network} (MLCNN) to perform multivariate time series forecasting. It first leverages a Convolution Neural Network (CNN)~\cite{CNN} to extract multi-level feature abstractions from the raw time series and engages them for multiple predictive tasks. Next, the extracted abstractions are fed into a shared Long Short-Term Memory (LSTM)~\cite{LSTM}, which captures complex long-term dependencies of the time series and fuses the future visions of different predictive tasks. In addition, we design another main LSTM for the main predictive task, utilizing the feature abstraction and the shared information of the primary task to make more accurate prediction. In this case, the shared LSTM works as an encoder for the features of the primary task and the main LSTM works as a decoder. Finally, similar to the method proposed by~\cite{LSTNet}, to deal with the scale changing problem of input data, we combine traditional autoregressive linear models with the non-linear part of neural networks to make our MLCNN more robust. Our contributions are of three-folds:
\begin{itemize}
\item Based on the Construal Level Theory about human predictive behavior, we design an effective extraction-sharing mechanism to construct and fuse the future visions of different forecasting tasks, and demonstrate its capabilities of improving the main predictive task.
\item We develop a novel multi-task deep learning model with good generalization ability for multivariate time series forecasting.
\item We conduct extensive experiments on three real-world datasets and show the advantages of our model against most state-of-the-art baseline methods, demonstrating new benchmark on the public datasets. All the data and experiment codes of our model are available at Github\footnote{\url{https://github.com/smallGum/MLCNN-Multivariate-Time-Series}}.
\end{itemize}

\section{Related Work}
\label{sec:related_work}

\textbf{Time Series Forecasting} Research on time series forecasting has a long history. One of the most popular models is the Autoregression (AR) model. The variants of AR model such as the Moving Average (MA), Autoregressive Integrated Moving Average (ARIMA), and Vector Autoregression (VAR) models are also widely used~\cite{TS02}. However, the AR model and its variants fall short in capturing the non-linear features of the time series signals due to their linear assumption about the data~\cite{TS08}. To address this problem, various non-linear models have been proposed, such as Factorization Machine (FM)~\cite{QoS02,QoS03,QoS01} and Support Vector Regression (SVR)~\cite{SVR}. Nevertheless, the number of parameters in these models grows quadratically over the temporal window size and the number of variables, implying large computational cost and high risk of overfitting when dealing with high dimensional multivariate time series.

Recently, Deep Neural Networks (DNNs) have attracted increasing attentions in the domain of time series forecasting, due to their great success in capturing non-linear data features. The first widely used models are Multi-Layer Perceptrons (MLPs)~\cite{TS05}, which learn the non-linear relationships of the input series through fully connected hidden layers. Later, Recurrent Neural Networks (RNNs) are known for their advantages in sequence learning~\cite{Seq2seq}. In order to solve the vanishing gradients problem~\cite{GDV} when using RNNs to learn long-term dependencies, the Long Short-Term Memory (LSTM)~\cite{LSTM} and the Attention~\cite{Attention} models have been proposed and achieved thrilling results on univariate time series forecasting with multiple driving series~\cite{GeoMAN,DARNN}. Besides, Convolutional Neural Networks (CNNs)~\cite{CNN} have also found their significance on asynchronous time series prediction~\cite{SOCNN}. Furthermore, both theoretical and empirical findings have suggested that combining autoregressive linear models with non-linear DNNs can effectively improve the predictive performance~\cite{Survey02,LSTNet}. Such a hybrid method is also adopted by our MLCNN model.

\textbf{Construal Level Theory} During the centuries, psychologists have conducted a large amount of researches on how individuals predict the future and the factors that influence those predictions~\cite{CLT02}. Particularly, Construal Level Theory (CLT) and its following study have been trying to reveal how temporal distance from future outcomes affect people's predictions~\cite{CLT01,CLT05}. CLT assumes that individuals' predictions of future events depend on how they mentally construe those events. According to CLT, people tend to use higher level, more abstract construals to represent distant future events than to represent near future events~\cite{CLT07}. For multivariate time series forecasting, CLT inspires us to extract more abstract representations of data for distant future predictions and more specific features for near future predictions.

CLT is the core of the proposed architecture in this paper. Our MLCNN model uses a multi-layer CNN to extract discriminative features of the raw time series at different convolutional layers, forming the \emph{construals} of different abstraction levels. The low- and high-level construals are respectively used for near and distant future predictions, thus producing near and distant future visions for the fusion model.

\textbf{Multi-Task Deep Learning} Multi-task learning (MTL) \cite{MTL01} aims to train multiple tasks in parallel, so as to improve the performance of the main task with training signals from other related tasks. MTL in deep neural networks, called multi-task deep learning, have achieved significant results in many areas of artificial intelligence~\cite{MTDL}. However, literature on multi-task deep learning for time series prediction is still scarce, mainly due to the difficulties of finding proper auxiliary tasks. \cite{AECRNN} proposed an MTL model AECRNN to perform univariate time series forecasting with related driving series, while few literature apply it on multivariate time series forecasting.

Our MLCNN model is a natural multi-task deep learning framework for multivariate time series forecasting. We choose the near and far future predictive tasks defined in Figure~\ref{fig:future_visions} as the auxiliary tasks and fuse their forecasting information to improve the main task. We demonstrate the superiority of this method through extensive experiments.

\section{Model Architecture}
\label{sec:model_arch}

In this section, we first formulate the problem at hand, and then present the proposed MLCNN architecture. Finally, we introduce the loss function and the optimization algorithm used by our model.

\subsection{Problem Statement}
\label{subsec:ps}

In this paper, we focus on the task of multivariate time series forecasting. More formally, given time series $\mathcal{X} = \{X_{t-p+1}, ..., X_{t}\}$, where $X_{i} \in \mathbb{R}^n$ and $n$ is the variable dimension, we are interested in predicting the value of the series at a certain future moment, that is, predicting the value of $X_{t+h}$, where $h \ge 1$ is the desirable horizon ahead of the current time stamp. In practice, the horizon $h$ is chosen according to the demands of the environmental settings.

Besides, we define two notations $fsp$, namely \emph{future span} and $fst$, namely \emph{future stride} to help specify auxiliary forecasting tasks, where $0 < fsp \cdot fst < h$. Therefore, while performing prediction on the series value at the future moment $t+h$ as the main task, we also perform predictions at future moments $\{t+h-(fsp \cdot fst), ..., t+h-fst, t+h+fst, ..., t+h+(fsp \cdot fst)\}$ as auxiliary tasks. Without loss of generality, we set $fsp=2$ and $fst=1$ by default. That being said, assuming $\{X_{t-p+1}, ..., X_{t}\}$ are available, we predict the values of $\{X_{t+h-2}, X_{t+h-1}, X_{t+h}, X_{t+h+1}, X_{t+h+2}\}$ in parallel, forming a five-task learning problem. Among the five tasks, predictions on $X_{t+h-2}$ and $X_{t+h-1}$ have near future visions while predictions on $X_{t+h+1}$ and $X_{t+h+2}$ have distant future visions.

\subsection{Convolutional Component}

\begin{figure*}[t]
\centering
\includegraphics[width=0.8\textwidth]{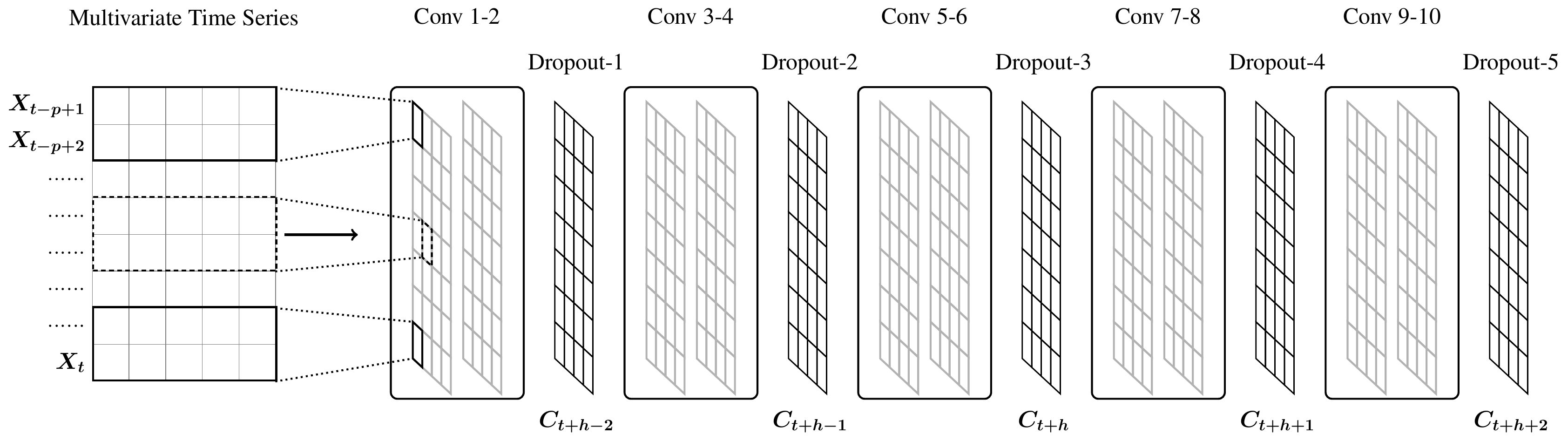} 
\caption{A 10-layer CNN to extract multi-level construals of the raw data}
\label{fig:CNN}
\end{figure*}

The first part of MLCNN is a multi-layer CNN~\cite{CNN}, where different layers extract different abstract features from the input data and deeper layers produce more abstract information. The CNN component aims to learn the local dependencies between variables and manufacture construals of different abstraction levels for multiple predictive tasks. As shown in Figure~\ref{fig:CNN}, for the five-task forecasting problem described above, we use the CNN to create five different construals:
\begin{align}
\begin{split}
  C_{t+h-2} &= f_1(\mathbf{X_t^{-p}}) \\
  C_{t+h-1} &= f_2(C_{t+h-2})  \\
  C_{t+h} &= f_3(C_{t+h-1}) \\
  C_{t+h+1} &= f_4(C_{t+h}) \\
  C_{t+h+2} &= f_5(C_{t+h+1}) ~,
\end{split}
\end{align}
where
\begin{itemize}
\item $\mathbf{X_t^{-p}} = [X_{t-p+1};X_{t-p+2};...;X_{t}] \in \mathbb{R}^{p \times n}$ is the matrix of the given multivariate time series. $n$ is the number of variables and $p$ denotes the number of time points.
\item $f_i : \begin{cases}
  \mathbb{R}^{p \times n} \to \mathbb{R}^{p \times m} & i = 1 \\
  \mathbb{R}^{p \times m} \to \mathbb{R}^{p \times m} & i = 2,3,4,5 
\end{cases}$ are one-dimensional convolutional layers (Conv1D) with $m$ filters in the CNN, and layer $f_{i+1}$ is deeper than layer $f_i$.
\item $C_{t+h-2}, ..., C_{t+h+2} \in \mathbb{R}^{p \times m}$ are extracted construals used for predictive tasks on $X_{t+h-2}, ..., X_{t+h+2}$, respectively. Dropout operation~\cite{Dropout} is also applied on every construal to avoid overfitting.
\end{itemize}

In addition, each filter in the CNN is $W_k \in \mathbb{R}^{w \times n}$ (the height of the filter is set to be the same as the variable dimension). The $k$-th filter sweeps through the input matrix $X$ and produces:
\begin{equation}
  c_k = Act(W_k \ast X + b_k) ~,
\end{equation}
where $\ast$ denotes the convolution operation and $c_k$ is the output vector. $Act$ could be any activation function. In this paper, we empirically found that the $LeakyReLU$ function $LeakyReLU(x) = \begin{cases}
x & x \ge 0 \\
\alpha x & \text{otherwise}
\end{cases}$
with leak rate $\alpha = .01$ fits most data well. We make each vector $c_k$ of length $p$ by zero-padding on the matrix $X$.

\begin{figure*}[t]
\centering
\includegraphics[width=0.8\textwidth]{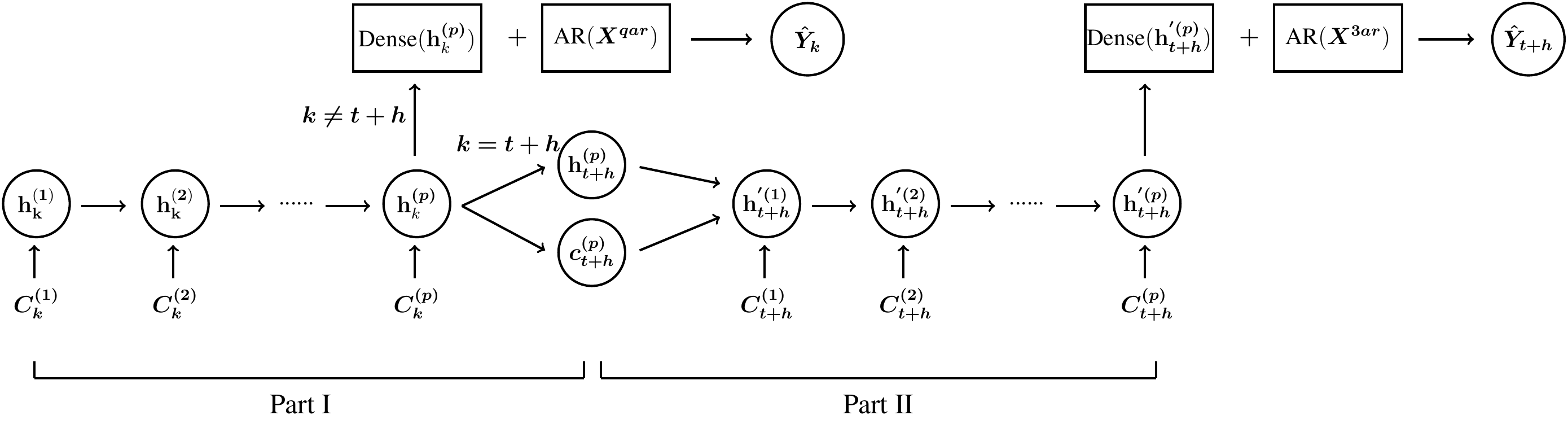} 
\caption{Part I: The shared LSTM for all tasks; Part II: The main LSTM for main task}
\label{fig:LSTM}
\end{figure*}

\subsection{Shared Recurrent Component}

The construals $C_{t+h-2}, ..., C_{t+h+2}$ of multiple abstraction levels are then fed into a shared RNN one after another. The recurrent component is an LSTM~\cite{LSTM} with the $\tanh$ function as the hidden update activation function. It captures the long-term dependencies of the time series and models the interactions between different predictive tasks, as shown in Figure~\ref{fig:LSTM} Part \uppercase\expandafter{\romannumeral1}. The hidden state of recurrent units at time $\tau$ for the $k$-th construal is computed as:
\begin{align}
\begin{split}
  i_k^{(\tau)} &= \sigma (W_{ii}C_k^{(\tau)} + b_{ii} + W_{hi} \mathbf{h}_k^{(\tau-1)} + b_{hi}) \\
  f_k^{(\tau)} &= \sigma (W_{if}C_k^{(\tau)} + b_{if} + W_{hf}\mathbf{h}_k^{(\tau-1)} + b_{hf}) \\
  g_k^{(\tau)} &= \tanh(W_{ig}C_k^{(\tau)} + b_{ig} + W_{hg}\mathbf{h}_k^{(\tau-1)} + b_{hg}) \\
  o_k^{(\tau)} &= \sigma (W_{io}C_k^{(\tau)} + b_{io} + W_{ho}\mathbf{h}_k^{(\tau-1)} + b_{ho}) \\
  c_k^{(\tau)} &= f_k^{(\tau)} \odot c_k^{(\tau-1)} + i_k^{(\tau)} \odot g_k^{(\tau)} \\
  \mathbf{h}_k^{(\tau)} &= o_k^{(\tau)} \odot \tanh(c_k^{(\tau)}) ~,
\end{split}
\end{align}
where $k \in \{t+h-2,t+h-1,...,t+h+2\}$, $1 \le \tau \le p$, $C_k^{(\tau)}$ denotes the $\tau$-th row of the construal $C_k$, $\odot$ is the element-wise product and $\sigma$ is the sigmoid function. The initial hidden state $\mathbf{h}_k^{(0)}$ and the initial cell state $c_k^{(0)}$ are set to zero by default. This shared LSTM fuses all kinds of future visions by sharing its weights and biases across all predictive tasks. Therefore, the fusion information is stored in shared parameters after the training phase and produces fusion features $\mathbf{h}_{t+h-2}^{(p)}, \mathbf{h}_{t+h-1}^{(p)}, \mathbf{h}_{t+h}^{(p)}, \mathbf{h}_{t+h+1}^{(p)}, \mathbf{h}_{t+h+2}^{(p)}$ for each forecasting task during the testing phase.

\subsection{Main Recurrent Component}

Similar in spirit to the Encoder-Decoder architecture~\cite{Encoder}, for the main predictive task (i.e. the forecasting of $X_{t+h}$), we use the shared LSTM to encode its fusion sequence, and devise another main LSTM to predict the output sequence. As shown in Figure~\ref{fig:LSTM} Part \uppercase\expandafter{\romannumeral2}, the output sequence $\mathbf{h}_{t+h}^{'(p)}$ is computed as:
\begin{equation}
  \mathbf{h}_{t+h}^{'(p)} = MainLSTM(C_{t+h}, \mathbf{h}_{t+h}^{(p)}, c_{t+h}^{(p)}) ~,
\end{equation}
where $MainLSTM$ has the same architecture as the shared LSTM but with the initial hidden state and cell state set to $\mathbf{h}_{t+h}^{(p)}$ and $c_{t+h}^{(p)}$, respectively. In our experiments, we empirically found that such a Fusion-Encoder-Main-Decoder architecture can boost the model performance in most cases.

We use a dense layer to align the outputs of the shared and main LSTMs: 
\begin{equation}
  r^D_k = \begin{cases}
  W_k^D\mathbf{h}_{k}^{'(p)} + b_k^D & k = t+h \\
  W_k^D\mathbf{h}_{k}^{(p)} + b_k^D & \text{otherwise}
  \end{cases} ~,
\end{equation}
where $k \in \{t+h-2,...,t+h+2\}$, $r^D_k \in \mathbb{R}^{n}$ is the prediction result of the neural network for the predictive task on $X_k$ and $W_k^D, b_k^D$ are weights and biases of the dense layer.

\subsection{Autoregressive Component}

As is pointed out by~\cite{LSTNet}, the non-linear nature of CNN and LSTM leads to their poor performance in capturing the scale changes of inputs, which significantly lowers the forecasting accuracy of the neural network model. To address this deficiency, Lai et al. decomposes their model into a linear part (i.e. the Autoregressive model) and a non-linear part (i.e. the neural network model). In this paper, we adopt the same method but change the Autoregressive (AR) model architecture to fit the neural network component of the MLCNN model. Typically, the scale of near future values is sensitive to the scale of recent past values, while the scale of distant future values is sensitive to the scale of both recent and further past values. Hence, we denote $s^{ar} \in \mathbb{N}$ as the \emph{autoregressive stride} and define $X^{qar} = [X_t;X_{t-1};...;X_{t-qs^{ar}+1}] \in \mathbb{R}^{qs^{ar} \times n}$, where $q \in \{1,2,3,4,5\}$. The forecasting result of the AR component for each predictive task is computed as follows:
\begin{equation}
  r^L_{k,i} = \sum_{j=0}^{qs^{ar}} W^{L}_{k,j} X^{qar}_{j,i} + b^{L}_k ~,
\end{equation}
where $k = t+h-3+q, 1 \le i \le n$. $r^L_{k,i}$ and $X^{qar}_{j,i}$ denote the $i$-th element of vectors $r^L_k$ and $X^{qar}_{j}$, respectively. And $W^{L}_k, b^{L}_k$ are weights and biases of the AR model. Note that all dimensions share the same set of linear parameters in each task.

We obtain the final prediction by combining the result of the neural network component and the AR component:
\begin{equation}
  \hat Y_k = r^D_k + r^L_k ~,
\end{equation}
where $k \in \{t+h-2,...,t+h+2\}$ and $\hat Y_k \in \mathbb{R}^{n}$ is the final prediction of $X_{k}$. Thus $\hat Y \in \mathbb{R}^{5 \times n}$ is the final prediction matrix of the five-task learning problem described previously.

\subsection{Loss Function}

$L_2$ error is the default loss function for most of the time series forecasting tasks:
\begin{equation}
  L_2(Y,\hat Y) = \sum_{\Omega_{\text{train}}} \sum_{k=1}^l \sum_{j=1}^n (Y_{k,j} - \hat Y_{k,j})^2 ~,
\end{equation}
where $l$ is the number of tasks, $n$ is the number of variables, $Y$ is the ground truth, $\hat Y$ is the model's prediction and $\Omega_{\text{train}}$ denotes the training set. However, researchers have found that the absolute loss ($L_1$-loss) function:
\begin{equation}
  L_1(Y,\hat Y) = \sum_{\Omega_{\text{train}}} \sum_{k=1}^l \sum_{j=1}^n |Y_{k,j} - \hat Y_{k,j}|
\end{equation}
works better than $L_2$ loss function in some cases. In the experiment part of this paper, we use the validation set to decide which of the two loss functions is better for our model. The goal of training is to minimize the loss function given the parameter set of our model, which can be achieved by using the Stochastic Gradient Decent (SGD) method or its variants. In this paper, We utilize the Adam~\cite{adam} algorithm to optimize the parameters of our model.

\section{Experiments}
\label{sec:experiments}

In this section, we conduct extensive experiments on three real-world datasets for multivariate time series forecasting, and compare the result of proposed MLCNN model against 5 baselines. To demonstrate the efficiency of our model, we also perform time complexity analysis and ablation study.

\subsection{Datasets}

As depicted in Table~\ref{tab:data}, our experiments are based on three publicly available datasets: 
\begin{itemize}
\item Traffic~\cite{LSTNet}: This dataset consists of 48 months (2015-2016) \emph{hourly data} from the California Department of Transportation. It describes the road occupancy rates (between 0 and 1) measured by different sensors on San Francisco Bay area freeways.
\item Energy~\cite{Energy}: This UCI appliances energy dataset contains measurements of 29 different quantities related to appliances energy consumption in a single house, recorded \emph{every 10 minutes} for 4.5 months. We select 26 relevant attributes for our experiments.
\item NASDAQ~\cite{DARNN}: This dataset includes the stock prices of 81 major corporations and the index value of NASDAQ 100, collected \emph{minute-by-minute} for 105 days.
\end{itemize}

\begin{table}
  \centering
  \caption{Dataset statistics}
  \label{tab:data}
  \small
  \begin{tabular}{c||c|c|c}
    \toprule
    Dataset          & Traffic          & Energy          & NASDAQ          \\
    \toprule
    \#Instances      & 17544            & 19735           & 40560           \\
    \midrule
    \#Attributes     & 862              & 26              & 82              \\
    \midrule
    Sample rate      & 1 h              & 10 min          & 1 min           \\
    \midrule
    Train size       & 60\%             & 80\%            & 90\%            \\
    \midrule
    Valid size       & 20\%             & 10\%            & 5\%             \\
    \midrule
    Test size        & 20\%             & 10\%            & 5\%             \\
    \bottomrule
  \end{tabular}
\end{table}

\subsection{Metrics}

To evaluate the performance of different methods for multivariate time series prediction, we use two conventional evaluation metrics (1) Root Mean Squared Error: $RMSE = \sqrt{\frac{1}{n} \sum_{j=1}^n (Y_{t+h, j} - \hat Y_{t+h,j})^2}$ and (2) Mean Absolute Error: $MAE = \frac{1}{n} \sum_{j=1}^n |Y_{t+h, j} - \hat Y_{t+h,j}|$, where $n$ is variable dimension, $Y_{t+h} \in \mathbb{R}^n$ is the ground truth of the time series value at the future moment $t+h$ and $\hat Y_{t+h} \in \mathbb{R}^n$ is the model's prediction. For both metrics, lower value is better. Note that for our multi-task forecasting model, we only use RMSE and MAE of the \emph{main predictive task} for evaluation.

\begin{table*}
  \centering
  \caption{Forecasting results of all methods over the three datasets (best results displayed in \textbf{boldface})}
  \label{tab:result}
  \resizebox{0.95\textwidth}{!}{
  \begin{tabular}{ll|ccc|ccc|ccc}
    \toprule[1pt]
    Dataset               &           &                           & Traffic                   &                           &                           & Energy                    &                           &                           & NASDAQ                    &                           \\
    \midrule[.5pt]
                          &           &                           \multicolumn{3}{c|}{Future}                             &                           \multicolumn{3}{c|}{Future}                             &                           \multicolumn{3}{c}{Future}                              \\
    \midrule[.5pt]
    Metrics               & Models    & t+3                       & t+6                       & t+12                      & t+3                       & t+6                       & t+12                      & t+3                       & t+6                       & t+12                      \\
    \midrule[.5pt]                                                                                                                                                                                                                                       
    \multirow{5}{*}{RMSE} & VAR       & 0.0370$\pm$9E-04          & 0.0373$\pm$4E-04          & 0.0364$\pm$6E-04          & 15.514$\pm$0.002          & 16.253$\pm$0.007          & 16.950$\pm$0.004          & 2.725$\pm$0.104           & 3.049$\pm$0.143           & 3.048$\pm$0.043           \\
                          & RNN-LSTM  & 0.0298$\pm$5E-05          & 0.0304$\pm$2E-04          & 0.0299$\pm$2E-04          & 15.820$\pm$0.002          & 16.758$\pm$0.095          & 17.289$\pm$0.027          & 4.529$\pm$0.314           & 4.946$\pm$0.181           & 5.353$\pm$0.141            \\
                          & MTCNN     & 0.0295$\pm$3E-04          & 0.0297$\pm$2E-04          & 0.0304$\pm$2E-04          & 15.841$\pm$0.154          & 16.549$\pm$0.028          & 17.481$\pm$0.203          & 4.197$\pm$0.174           & 3.928$\pm$0.217           & 4.341$\pm$0.327            \\
                          & AECRNN    & 0.0286$\pm$2E-04          & 0.0291$\pm$3E-04          & 0.0295$\pm$3E-04          & 15.705$\pm$0.124          & 16.259$\pm$0.152          & 17.173$\pm$0.165          & 9.785$\pm$0.438           & 9.893$\pm$0.304           & 9.727$\pm$0.351            \\
                          & LSTNet    & 0.0269$\pm$1E-04          & 0.0278$\pm$3E-04          & 0.0280$\pm$2E-04          & 15.506$\pm$0.049          & \textbf{15.795$\pm$0.074} & 16.890$\pm$0.105          & 0.366$\pm$0.006           & 0.522$\pm$0.010           & 0.754$\pm$0.022           \\
    \cmidrule(lr){2-11}                                                                                                                                                                                                                                                                   
                          & MLCNN     & \textbf{0.0258$\pm$1E-04} & \textbf{0.0264$\pm$1E-04} & \textbf{0.0267$\pm$8E-05} & \textbf{15.130$\pm$0.087} & 15.994$\pm$0.047          & \textbf{16.782$\pm$0.125} & \textbf{0.365$\pm$0.002}  & \textbf{0.516$\pm$0.003}  & \textbf{0.739$\pm$0.004}  \\
    \midrule[.5pt]                                                                                                                                                                                                                                                                      
    \multirow{5}{*}{MAE}  & VAR       & 0.0255$\pm$1E-03          & 0.0256$\pm$4E-04          & 0.0246$\pm$7E-04          & 2.898$\pm$0.001           & 3.321$\pm$0.026           & 3.872$\pm$0.009           & 1.834$\pm$0.069           & 2.075$\pm$0.072           & 2.008$\pm$0.015           \\
                          & RNN-LSTM  & 0.0134$\pm$6E-05          & 0.0138$\pm$1E-04          & 0.0136$\pm$2E-04          & 2.733$\pm$0.072           & 3.049$\pm$0.079           & 3.668$\pm$0.128           & 2.205$\pm$0.091           & 2.344$\pm$0.051           & 2.452$\pm$0.064            \\
                          & MTCNN     & 0.0135$\pm$3E-04          & 0.0139$\pm$2E-04          & 0.0143$\pm$4E-04          & 3.415$\pm$0.098           & 3.896$\pm$0.062           & 4.312$\pm$0.131           & 2.433$\pm$0.038           & 2.375$\pm$0.081           & 2.397$\pm$0.070            \\
                          & AECRNN    & 0.0121$\pm$1E-04          & 0.0124$\pm$2E-04          & 0.0131$\pm$2E-04          & 2.269$\pm$0.078           & 3.013$\pm$0.072           & 3.395$\pm$0.057           & 4.370$\pm$0.361           & 4.500$\pm$0.267           & 4.370$\pm$0.350            \\
                          & LSTNet    & 0.0116$\pm$8E-05          & 0.0123$\pm$3E-04          & 0.0124$\pm$2E-04          & \textbf{1.795$\pm$0.014}  & 2.386$\pm$0.030           & 3.112$\pm$0.043           & 0.093$\pm$0.003           & 0.135$\pm$0.006           & 0.195$\pm$0.012           \\
    \cmidrule(lr){2-11}
                          & MLCNN     & \textbf{0.0110$\pm$1E-04} & \textbf{0.0113$\pm$1E-04} & \textbf{0.0114$\pm$8E-05} & 1.879$\pm$0.033           & \textbf{2.378$\pm$0.017}  & \textbf{3.036$\pm$0.044}  & \textbf{0.091$\pm$0.001}  & \textbf{0.130$\pm$0.001}  & \textbf{0.186$\pm$0.001}  \\                                                                                                                                                                                                                                                                    
    \bottomrule[1pt]
  \end{tabular}
  }
\end{table*}

\subsection{Baselines}

We compare the MLCNN model with 5 baselines as follows:
\begin{itemize}
\item \textbf{VAR}~\cite{TS02,TS01}: The well-known Vector Autoregression model for multivariate time series forecasting.
\item \textbf{RNN-LSTM}~\cite{LSTM}: The Recurrent Neural Network using LSTM cell. We combine an LSTM layer and a dense layer to perform multivariate time series forecasting.
\item \textbf{MTCNN}~\cite{CNN}: The classical Convolution Neural Network exploiting the same multi-task idea as the MLCNN model. We use a simple multi-layer CNN to perform multiple predictive tasks of different future time.
\item \textbf{AECRNN}~\cite{AECRNN}: A multi-task learning model combining additional auto-encoders with a unified framework of Convolution and Recurrent Neural Networks. AECRNN is originally designed to perform univariate time series forecasting given other correlated time series. Here we extend this model to perform multivariate time series forecasting and compare it with the multi-task learning framework proposed in this paper.
\item \textbf{LSTNet}~\cite{LSTNet}: A novel multivariate time series forecasting framework which achieves great performance improvements by catching the long-term and short-term patterns of the time series data.
\end{itemize}

\subsection{Training Details}

We conduct a grid search over all hyperparameters for each method and dataset. Specifically, for the length of input window size $T$, we set $T \in \{1, 6, 12, 24, 24 \times 7\}$-hour and all methods share the same grid search range. For RNN-LSTM, we vary the number of hidden state size in $\{10, 25, 50, 100, 200\}$. For MTCNN, the filter number of CNN is chosen from $\{5, 10, 25, 50, 100\}$. For AECRNN, we use the default settings by its author. For LSTNet and MLCNN, the filter number of CNN is chosen from $\{5, 10, 25, 50, 100\}$ and the hidden state size of RNN is chosen from $\{10, 25, 50, 100, 200\}$. For simplicity, we use the same hidden state size for shared LSTM and main LSTM of our MLCNN model. The dropout rate of our model is chosen from $\{0.2, 0.3, 0.5\}$. During the training phase, the batch size is $128$ and the learning rate is $0.001$. We test different hyperparameters and find the best settings for each method.

\subsection{Main Results}

Table~\ref{tab:result} summaries the experimental results of all the methods on the three datasets. Following the testing settings of~\cite{LSTNet} and~\cite{DARNN}, we use each model to predict the future value of the time series at the future moment $\{t+3, t+6, t+12\}$, respectively, which means the future moment is set from 3 to 12 hours for the forecasting over the Traffic data, from 30 to 120 minutes over the Energy data and from 3 to 12 minutes over the NASDAQ data. To be fair, we train each model under different parameter settings for 10 times and report their best average performance and standard deviations for comparison\footnote{We first use the validation set to select hyper-parameters that obtain similar good predictive results. And then we simply split the dataset into training and testing sets to retrain the model and show the best testing performance of those good hyper-parameters.}.

\begin{figure}[t]
  \centering
  \includegraphics[width=.95\columnwidth]{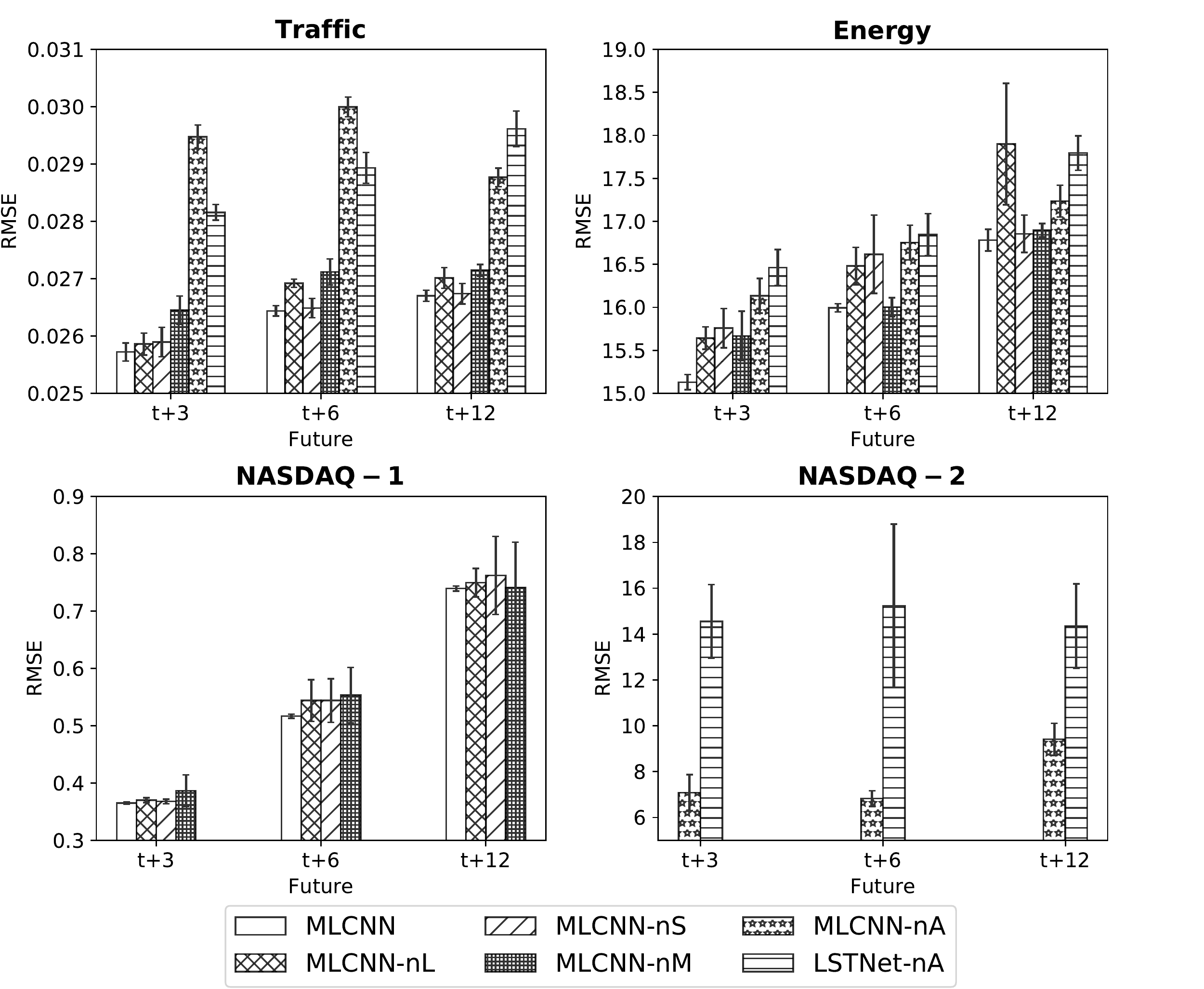}
  \caption{Results of variant comparison}
  \label{fig:comp}
\end{figure}

Clearly, our method outperforms most of the baselines on both metrics, demonstrating the effectiveness of the framework design for fusing different future visions. The most significant improvements are achieved on the Traffic dataset. Specifically, MLCNN outperforms the state-of-the-art baseline LSTNet by \textbf{4.09\%}, \textbf{5.04\%}, \textbf{4.64\%} on RMSE and \textbf{5.17\%}, \textbf{8.13\%}, \textbf{7.32\%} on MAE on the Traffic dataset. Remarkably, our model achieves significantly improvements beyond the state-of-the-art multi-task learning framework AECRNN over all of the three datasets, showing the advantages of utilizing multi-level construals for multi-task forecasting. On the other hand, we observe that the RMSE and MAE of RNN-LSTM, MTCNN and AECRNN models are much worse than other models on the NASDAQ dataset. This is mainly because the three models are not sensitive to the scale of the input data due to their lack of the AR component. Therefore, the non-periodic scale changing of input signals will dramatically lower their predictive performance. In Figure~\ref{fig:comp}, we also show the failure of MLCNN and LSTNet on NASDAQ dataset without the AR component. Furthermore, we conduct the two-sample $t$-test~\cite{ttest2} at 5\% significance level on the forecasting results of MLCNN and LSTNet models. Overall, the small $p$-value of the test shows that our model achieves statistically significant improvements beyond the LSTNet model. 

In summary, the evaluation results demonstrate the success of our MLCNN framework in fusing near and distant future visions to improve the forecasting accuracy.

\begin{figure}[t]
  \centering
  \includegraphics[width=.95\columnwidth]{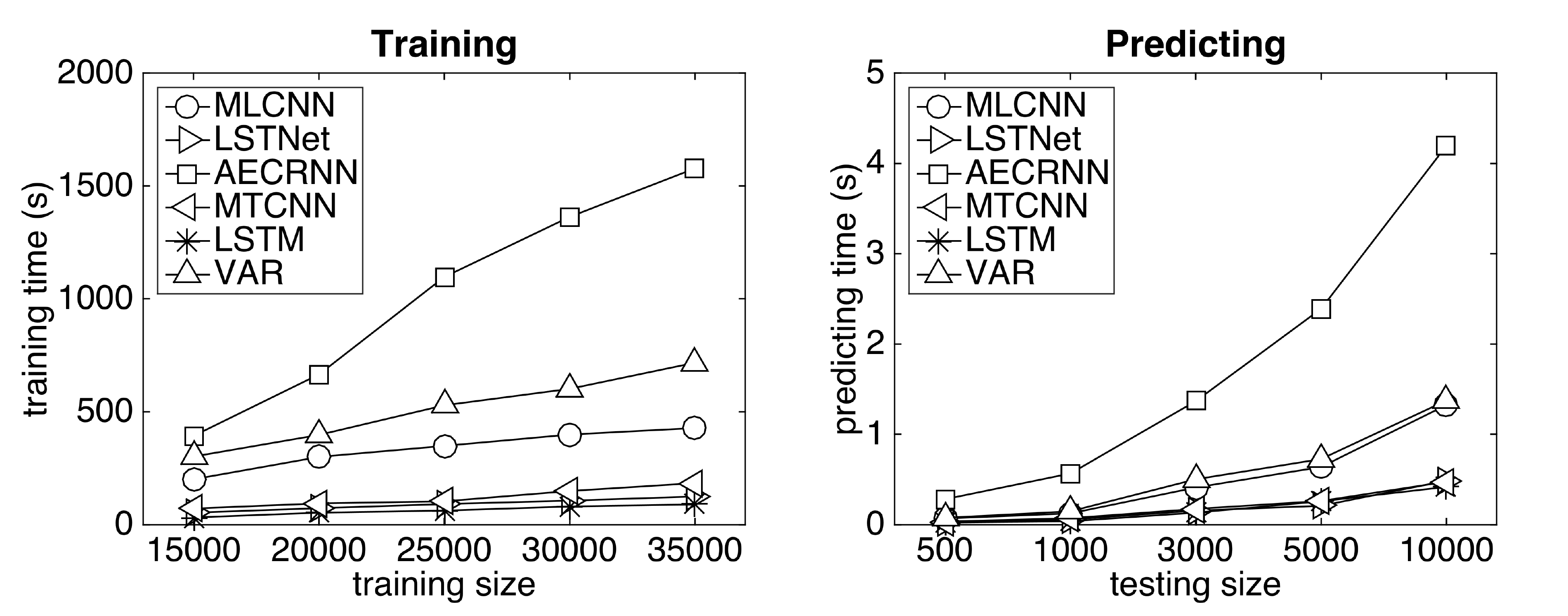}
  \caption{Result of time complexity comparison}
  \label{fig:training_speed}
\end{figure}

\subsection{Time Complexity Analysis}

\noindent Although the proposed MLCNN architecture appears a little complex, we believe that the sharing mechanism of multi-task learning helps to reduce the training and predicting complexity. In the convolutional component, all preditive tasks share the low layers of a single multi-layer CNN. Also, in the recurrent component, weights and biases of the fusion LSTM are shared by all tasks. Sharing parameters among different tasks ensures that model complexity will not increase too much while performing multiple tasks. To prove this, we compare the behavior of all models as a function of the sample size and show the result over the NASDAQ dataset in Figure~\ref{fig:training_speed}. The training and predicting time of our MLCNN model is close to that of other baselines. Significantly, MLCNN outperforms the VAR and the AECRNN models when dealing with high dimensional time series, proving the efficiency of our multi-task learning design.

\subsection{Variant Comparison}
\label{subsec:vc}

To demonstrate the effectiveness of each model component, we compare MLCNN with 5 variants as follows:
\begin{itemize}
\item \textbf{MLCNN-nL}: We remove differences of abstraction levels between the construals for multiple tasks. Instead, we use independent CNNs with the same number of convolutional layers to construct each construal and thus fail the CLT in our model.
\item \textbf{MLCNN-nS}: We remove the shared LSTM component (i.e., the fusion encoder) such that there is no fusion for the future visions of different forecasting tasks.
\item \textbf{MLCNN-nM}: We remove the main LSTM component (i.e., the main decoder) and use a single LSTM for fusion and prediction.
\item \textbf{MLCNN-nA}: We remove the AR component and test the predictive performance of the neural network part.
\item \textbf{LSTNet-nA}: We also remove the AR component of the LSTNet model and compare it with MLCNN-nA.
\end{itemize}
For all the variants, we tune their hidden dimension to make them have similar numbers of model parameters to the completed MLCNN model, eliminating the influences of different model complexity.

Figure~\ref{fig:comp} presents the results of comparison. Important observations from these results are listed as follows:
\begin{itemize}
\item MLCNN achieves the best result on all the datasets.
\item Removing any component from MLCNN not only causes the performance drops but also increases the variances, showing the robustness of our MLCNN architecture design.
\item Removing the AR component (in MLCNN-nA) from MLCNN causes the most significant performance drops on most of the datasets, which verifies the scale insensitive problem proposed by~\cite{LSTNet}.
\item MLCNN-nA achieves better performance than LSTNet-nA on most of the datasets, demonstrating the advantages of the neural network component of our MLCNN model even without the AR component.
\end{itemize}
In conclusion, the full MLCNN architecture is the most effective and robust forecasting model across all experiment settings.

\begin{figure}[t]
  \centering
  \includegraphics[width=.95\columnwidth]{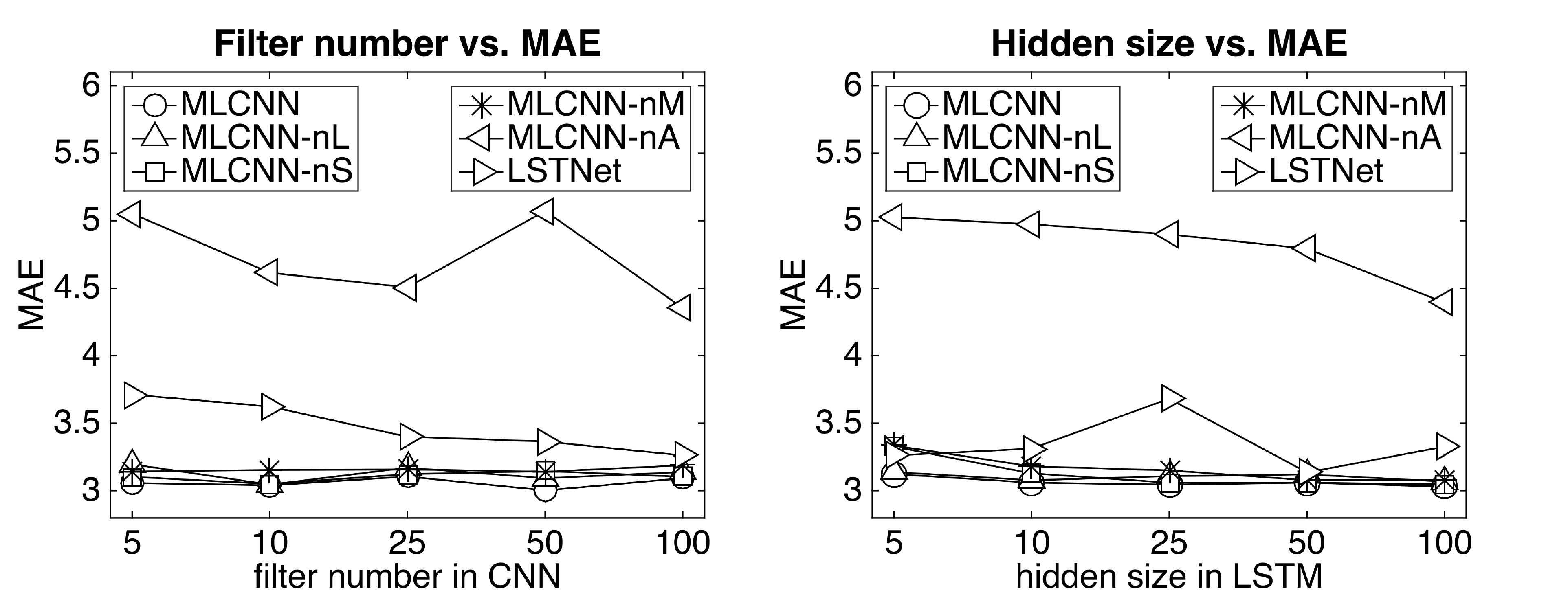}
  \caption{Results of parameter sensitivity tests}
  \label{fig:para_sen}
\end{figure}

Furthermore, we try different filter number of CNN and hidden state size of LSTM in both MLCNN as well as its variants and LSTNet. Figure~\ref{fig:para_sen} shows the comparison results of prediction on $X_{t+12}$ on the Energy dataset. We can observe that MLCNN generally achieves best results under different parameter settings. Besides, compared to the LSTNet and the variants, our model is less sensitive to the parameter changes, showing the effectiveness of our multi-task deep learning framework.

\section{Conclusion}
\label{sec:conclusion}

In this paper, we propose a novel multi-task deep learning framework (MLCNN) for multivariate time series forecasting. In the first level, based on the Construal Level Theory of psychology, we design a multi-layer Convolution Neural Network to produce multi-level abstract construals for multiple predictive tasks. In the second level, we devise a Fusion-Encoder-Main-Decoder architecture to fuse the future visions of all tasks. Moreover, we combine the autoregressive model with the neural network to boost predictive performance. Experiments on three real-world datasets show that our model achieves the best performance against 5 baselines in terms of the two metrics (RMSE and MAE). In addition, we demonstrate the efficiency and robustness of the MLCNN architecture through in-depth analysis.

For the future research, the proposed model can be extended further by adding weighting machanism to the fusion encoder of different future visions, such as the Attention machanism~\cite{Attention}. Besides, how to dynamically choose the temporal distances from the future (i.e., the $fsp$ and $fst$ parameters) instead of setting their values to default is another challenging problem.

\section{Acknowledgments}
This paper was supported by the National Key Research and Development Program (2016YFB1000101), the National Natural Science Foundation of China (61722214, U1811462, 61876155), the Guangdong Province Universities and Colleges Pearl River Scholar Funded Scheme (2016) and Key Program Special Fund in XJTLU under no. KSF-A-01, KSF-E-26 and KSF-P-02.

\bibliography{AAAI-ChengJ.2179}
\bibliographystyle{aaai}
\end{document}